\documentclass[runningheads]{llncs}
\usepackage{graphicx, subfig, multirow}
\usepackage{cite}
\usepackage{svg}
\usepackage{hyperref}
\usepackage{verbatim}
\begin{document}
\title{Development and evaluation of intraoperative ultrasound segmentation with negative image frames and multiple observer labels}
%
\titlerunning{Classifier-assisted Segmentation}
%


\author{Liam F Chalcroft\inst{1, 2}\thanks{These authors contributed equally.}, Jiongqi Qu\inst{1}$^*$, Sophie A Martin\inst{1,3,4}$^*$, Iani JMB  Gayo\inst{1,3,5}*, Giulio V Minore\inst{6}*, Imraj RD Singh\inst{7}*, Shaheer U Saeed\inst{1,3,5}, Qianye Yang\inst{1,3,5}, Zachary MC Baum\inst{1,3,5}, Andre Altmann\inst{1,3}, Yipeng Hu\inst{1,3,5} \\
\email{liam.chalcroft.20@ucl.ac.uk}}
%

\institute{Department of Medical Physics and Biomedical Engineering, \and
Wellcome Center for Human Neuroimaging, \and
Centre for Medical Image Computing, \and
Dementia Research Centre, UCL Institute of Neurology, \and
Wellcome/EPSRC Centre for Interventional and Surgical Sciences, \and
Department of Physics and Astronomy, \and
Department of Computer Science, \\ 
University College London, London, UK\\}

\maketitle
\begin{abstract}
    When developing deep neural networks for segmenting intraoperative ultrasound images, several practical issues are encountered frequently, such as the presence of ultrasound frames that do not contain regions of interest and the high variance in ground-truth labels. In this study, we evaluate the utility of a pre-screening classification network prior to the segmentation network. Experimental results demonstrate that such a classifier, minimising frame classification errors, was able to directly impact the number of false positive and false negative frames. Importantly, the segmentation accuracy on the classifier-selected  frames, that would be segmented, remains comparable to or better than those from standalone segmentation networks. Interestingly, the efficacy of the pre-screening classifier was affected by the sampling methods for training labels from multiple observers, a seemingly independent problem. We show experimentally that a previously proposed approach, combining random sampling and consensus labels, may need to be adapted to perform well in our application. Furthermore, this work aims to share practical experience in developing a machine learning application that assists highly variable interventional imaging for prostate cancer patients, to present robust and reproducible open-source implementations, and to report a set of comprehensive results and analysis comparing these practical, yet important, options in a real-world clinical application.
\end{abstract}

\section{Introduction} 
Many urological procedures for prostate cancer patients, such as ablation therapy and needle biopsies, are guided by B-mode transrectal ultrasound images (TRUS) to identify and then monitor the shape and location of prostate glands \cite{Sarkar2016, Ghose2012}. This application is useful for a number of interventional tasks, such as estimating the gland size, regions of pathological interest and surrounding healthy, but vulnerable, tissues. However, due to variable acoustic coupling, inhomogeneous intensity distribution, and the necessity of real-time monitoring, delineating the boundaries of prostate glands is a challenging task, even for experienced urologists. Deep neural networks have been proposed to automate this process \cite{Hossain2019, Lei2019, Anas2018, Orlando2020, Wang2018}.

The performance of these networks relies on well-defined ground truth labels. To date, there is no gold standard approach in many ultrasound imaging applications with high inter- and intra-rater variability in labelling and its use in training. Existing approaches deal with multiple labels by using a pixel-level voting strategy or random sampling, both estimating the expected labels. In \cite{Sudre2019}, Sudre et. al observed that combining random and voting strategies during training improves stability and performance in the context of brain lesion detection. In this paper, we consider labels from multiple independent raters and investigate the effect of different sampling strategies during segmentation, and test these proposed sampling methods in the context of interventional ultrasound imaging for prostate cancer patients.

In addition to the label variability, ultrasound data itself is known to be of high variance, due to its user dependency and flexible use protocols. For example, it is common that some frames do not contain the region of interest (ROI), particularly due to the small size of the prostate gland in our application. The presence of negative frames presents a key challenge in segmentation, as wrongfully segmenting a frame that does not contain the ROI could potentially lead to misdiagnosis or damage to healthy tissues. Using a widely-used segmentation accuracy metric based on overlap, such as Dice, to quantify this error can be problematic. The naive implementation of Dice is independent of the number of false positive pixels and the cost of negative frames may not be easily quantified with respect to the cost of negative pixels when designing a new loss function. For example, in the case of a handheld setting, relative positions and distances in the out-of-plane direction between ultrasound frames are in general variable and unknown, which may lead to an unspecified misjudgement of where the ROI boundary is, given a false positive frame. A separate frame classification may provide more intuitive user guidance when using the segmentation algorithms. 

Limited work has been proposed to address the problem of negative frames within medical image segmentation. In \cite{Giordano}, false positives in a video object segmentation task were reduced through the introduction of a post-processing classifier.  In \cite{Rottman2019}, meta-classification was used to detect false positive samples in semantic segmentation. 
This has motivated a screening strategy in this work that can detect negative frames before they are incorrectly segmented by the segmentation network. Such a separately trained classification network can also provide flexible control at test-time between false positive and false negative rates on a frame-level, which is arguably more difficult to achieve by altering threshold on pixel-level class probability in a segmentation network. Alternative approaches and different loss functions to address this issue are also discussed or compared in this paper.

\section{Methods}

\subsection{Segmentation network}
U-Net \cite{unet_conv}, a fully convolutional neural network, is adapted from a well-established reference implementation. Our network consists of 5 layers that starting with initial 16 channels, with residual network blocks replacing the original individual convolutional layers to encourage fast convergence \cite{resnet}. Images were normalised to zero-mean and unit-variance. All the segmentation networks were trained with a mini-batch size of 32, using the Adam optimiser \cite{Kingma2017} with an exponential learning rate scheduler that minimises a soft Dice loss function:
$L_{Soft Dice} = \frac{2\Sigma y_{pred}\cdot y_{true}}{\Sigma y_{pred}+\Sigma y_{true}}$,
where $\Sigma$ is the pixel-wise sum, $y_{pred}$ is the predicted class probabilities and $y_{true}$ is the ground truth mask. The Dice value was also used to monitor validation set performance. Random data augmentations are applied during training with probability $p=0.3$, including random affine deformations (rotation $|\theta_r|\leq2.5\deg$, maximum translation $0.05$, scaling in range 0.95-1), and random flipping along the vertical axis. These augmentations were empirically found robust for the TRUS data in this application.

\subsection{Frame classification network}
A reference-quality ResNeXt \cite{Xie2017} classifier pre-trained on ImageNet was adapted to predict whether a prostate is present based on the frame-level consensus. The network was modified to accept single channel and resized to $224\times224$. The weights are normalised with mean and standard deviation (0.449, 0.226), representing the average of the three original RGB channels. This model was trained with an initial learning rate of 0.0001, using the Adam optimiser and a binary cross-entropy loss function.

\subsection{Label sampling}\label{label_sampling}
Six different label sampling methods were investigated and evaluation results on the hold-out test data are reported. The methods are summarised in Table \ref{table:label_methods}.  The combination label strategy randomly selects a certain percentage of the data to perform the vote sampling method and applies the random sampling method to the remainder data.

\setlength{\tabcolsep}{8pt}
\renewcommand{\arraystretch}{1.3}

\begin{table}[ht!]
    \caption{Summary of label sampling methods. Soft mean refers to the non-rounded mean of labels, treated as a continuous probability map.}\vspace{5pt}
    \label{table:label_methods}
\begin{tabular}{ |p{3cm} |p{8cm}| } 
\hline
Label strategy & Description \\
\hline
Vote & Pixel-level majority voting from the 3 labels \\
Random & Single label selected at random \\
Mean & Soft mean of the 3 labels \\
Combination ($25\%$) & Combination of 25\% vote and 75\% random labels \\
Combination ($50\%$) & Combination of 50\% vote and 50\% random labels  \\
Combination ($75\%$) & Combination of 75\% vote and 25\% random labels  \\
\hline 
\end{tabular}
\end{table}

\subsection{Pre-screening strategy}
The classifier can be combined with the segmentation network to facilitate a pre-screening strategy illustrated in Figure \ref{fig:pre-screening}. The frame will pass to the segmentation network only if the classifer-predicted probability is greater than a set threshold in logits, whose values from 0 to 5 are tested based on observations of resulting classification accuracy range on the validation set. Different ways of combining the classification and segmentation networks is also possible and remains interesting for future investigation. 

\subsection{Loss functions for segmentation}
For a given label sampling method, we test different segmentation loss functions. This allows us to ascertain whether the frame-level classification can also be handled by the segmentation directly,  as opposed to the above-described pre-screening. Two alternatives are considered in addition to the Dice loss function, a combo loss with an equal weighting between dice loss and binary cross entropy loss (BCE), and a weighted binary cross entropy loss based on \cite{Sudre2017} (W-BCE). The equation for the Dice-BCE loss is given by:
\begin{equation}
   Dice\mbox{-}BCE = 0.5 \times (1-Dice) + 0.5 \times BCE
\end{equation}
where the binary cross entropy loss is defined by:
\begin{equation}
    BCE = - \sum^{N} x_n \log{p_n} + (1-x_n) \log{(1-p_n)}
\end{equation} 
where N is the number of pixels, $x_n$ is target class per pixel and $p_n$ is the predicted probability from the network. The BCE loss can be modified to assign weights, $w_c$ to each class (c = 0, 1) such that: 
\begin{equation}
    W\mbox{-}BCE =  - \sum^{N} w_0 x_n \log{p_n} + w_1 (1-x_n) \log{(1-p_n)}
\end{equation}
where in our case, $w_1 = \frac{1}{\sum^{N} x_n = 1}$  and $w_0 = 1-w_1$.

\subsection{Evaluation experiments}
The Dice coefficient is computed on positive frames excluding those that are predicted to be negative by the segmentation network or by, when in use, the pre-screening classifer, to ensure that we do not penalise the network for correctly identifying negative frames (a 0 Dice coefficient). In addition, we report frame-level classification performance for both frame classifer and segmentation network, when the latter is used without pre-screening. In this case, rates of false positive frames and their false positive area are computed. All results are reported on the independent hold-out test set. \textit{p-values} from t-tests at significance level of 0.05 are also reported when comparison is made.

\begin{figure}[ht!]
\centering
\subfloat[ \label{fig:example_1}]{%
  \includegraphics[width=0.32\textwidth]{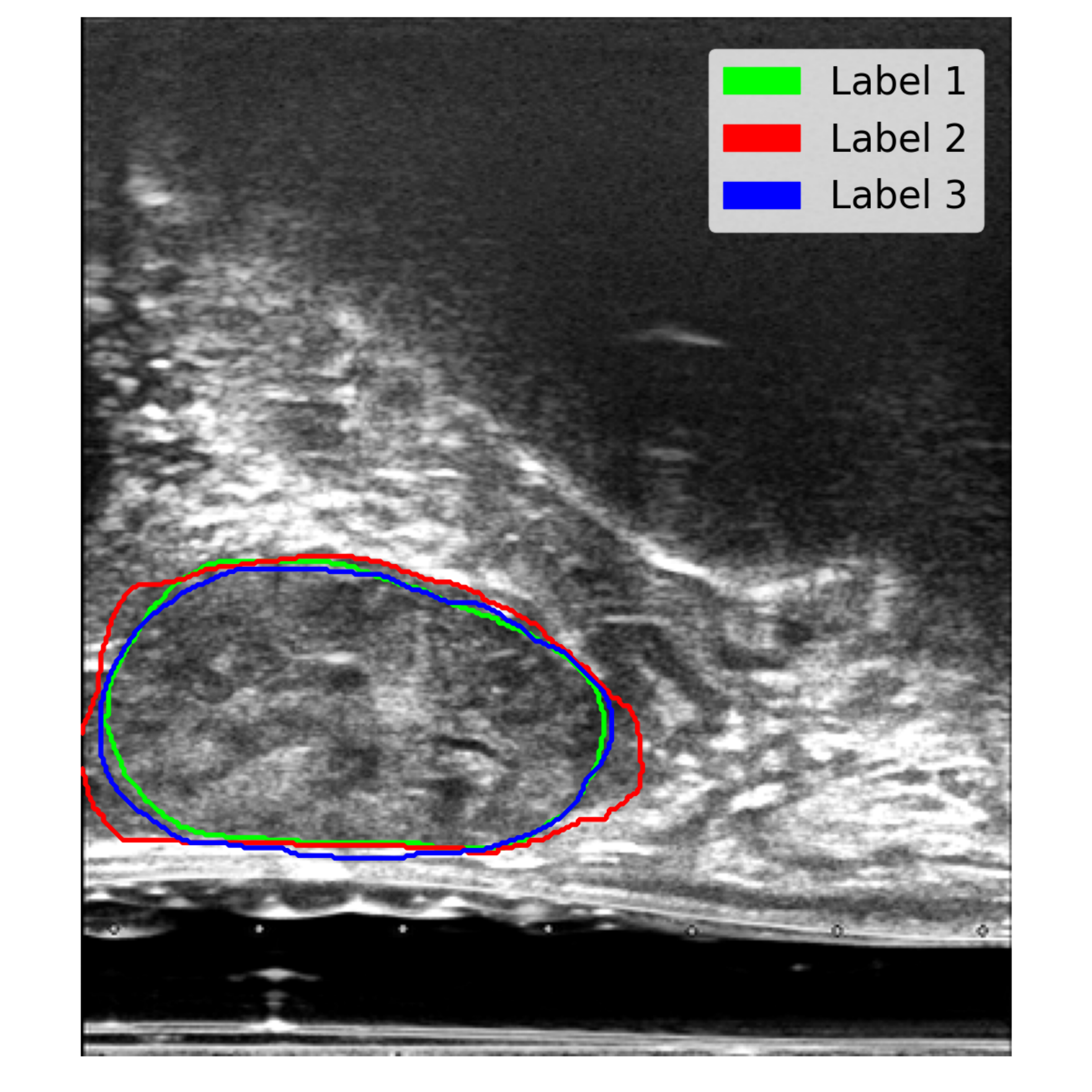}%
}\hfill
\subfloat[ \label{fig:example_2}]{%
  \includegraphics[width=0.32\textwidth]{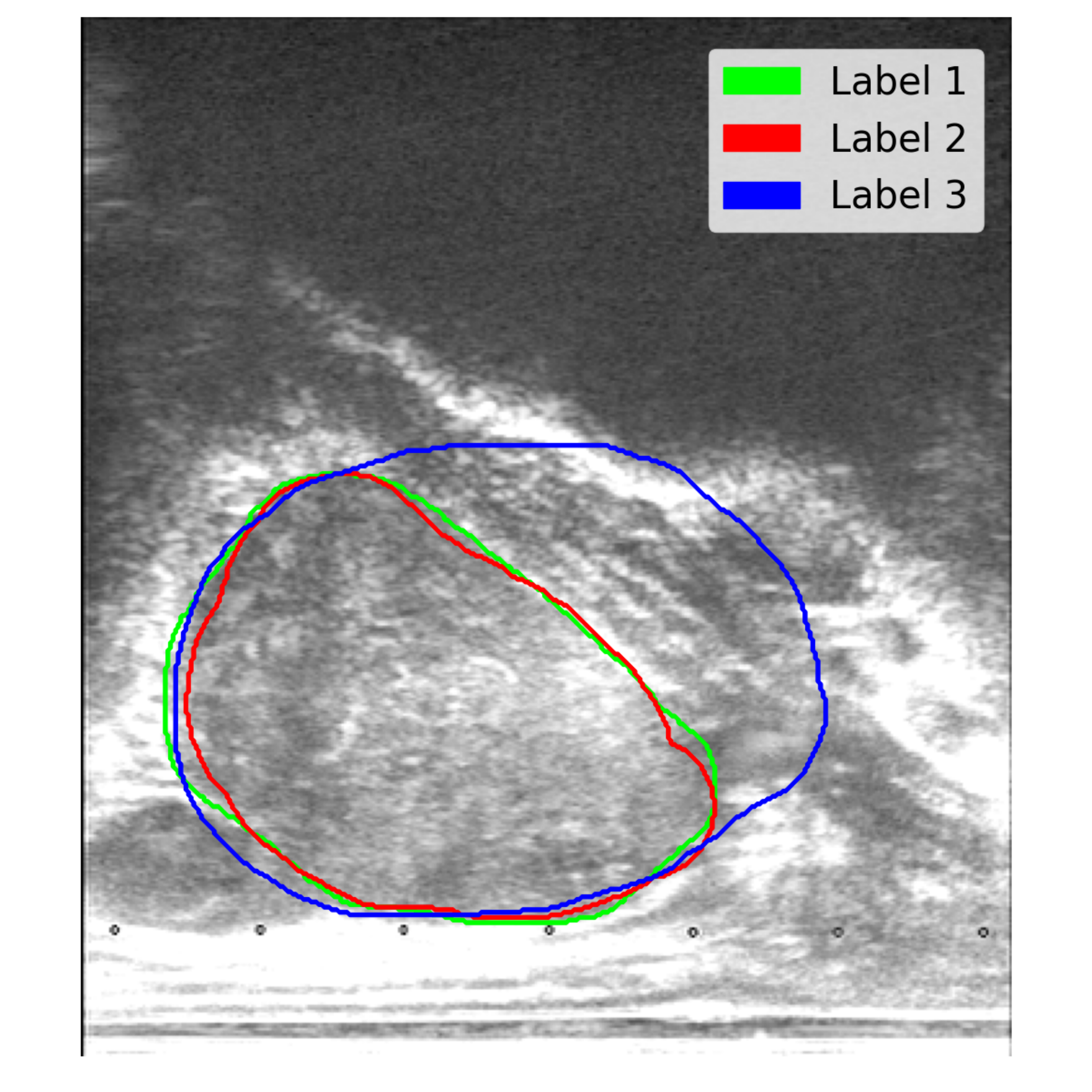}%
}
\subfloat[\label{fig:example_3}]{%
  \includegraphics[width=0.32\textwidth]{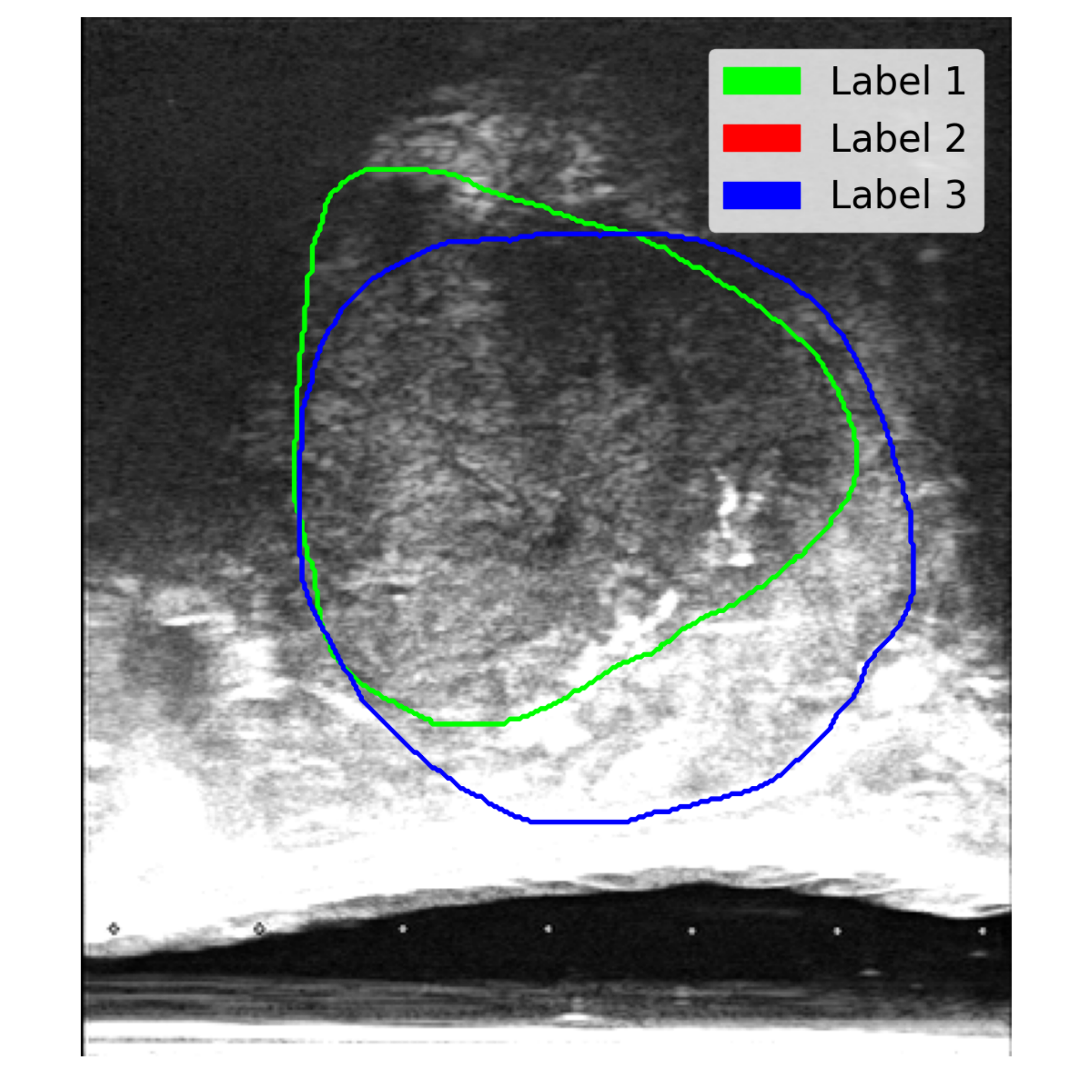}%
}\vfill
\subfloat[\label{fig:pre-screening}]{%
\includegraphics[width=1\textwidth]{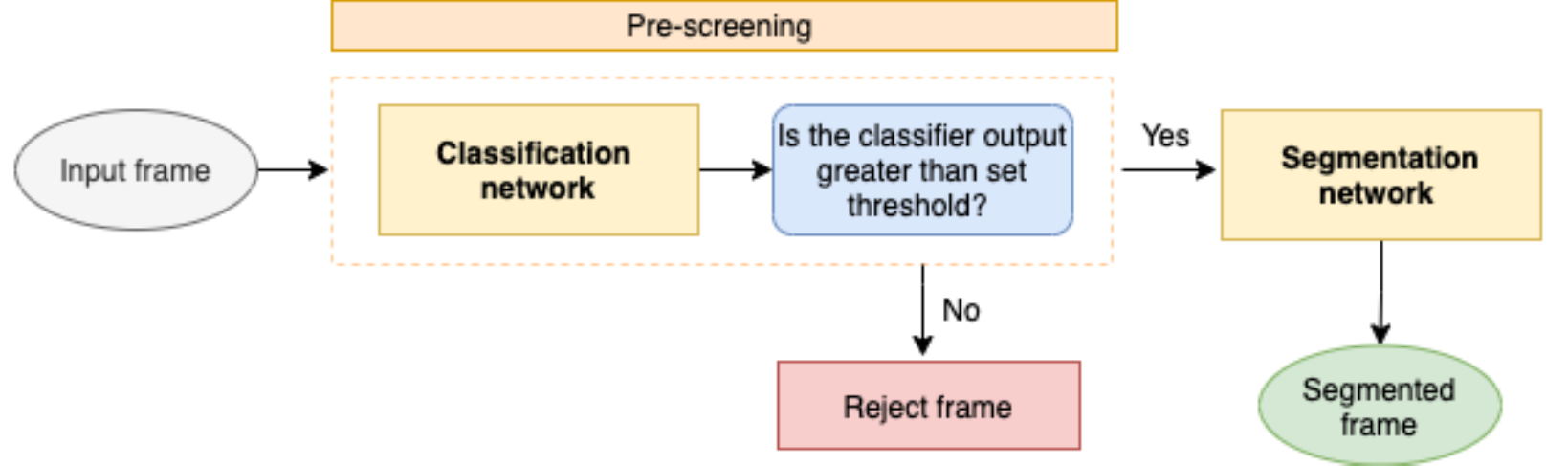}%
}
\caption{a-c) Example frames are shown with manual labels from three observers in green, red and blue respectively. a) All labels are in close agreement. b) Two labellers agree however one annotation is significantly larger. c) Only two labellers identify the prostate presence, but with slightly different locations. d) Flowchart to describe the pre-screening strategy.}
\label{fig:example_images}
\end{figure}

The dataset used in this study contains 2D B-mode transrectal ultrasound frames from 250 patients.
For each subject, a range of 50-120 frames were acquired at the start of the procedure, with a bi-plane transperineal ultrasound probe (C41L47RP, HI-VISION Preirus, Hitachi Medical Systems Europe) and a digital transperineal stepper (D\&K Technologies GmbH, Barum, Germany) to view and scan entire gland. For labelling, 6644 ultrasound images were sampled with size $403 \times 361$ and were manually annotated by three independent raters. A set of example frames are shown in Fig. \ref{fig:example_1}-\ref{fig:example_3} with varying label agreement.

At the patient level, 5224 and 1346 frames were sampled for training/validation and hold-out test, an 80:20 split. The networks were trained using a 3-fold cross-validation ensemble strategy, with 3484 and 1740 samples for training and validation in each fold, respectively. Predictions from each of the networks were averaged at test-time to generate a single probability map that is converted into a mask during inference on the hold-out set. The code is made publicly available at \href{https://github.com/sophmrtn/RectAngle}{https://github.com/sophmrtn/RectAngle}.

\section{Results and Discussion}
\subsubsection{Label sampling}
\begin{figure}[htp]
\centering
\subfloat[\label{fig:dice-sampling}]{\includegraphics[width=0.6\textwidth]{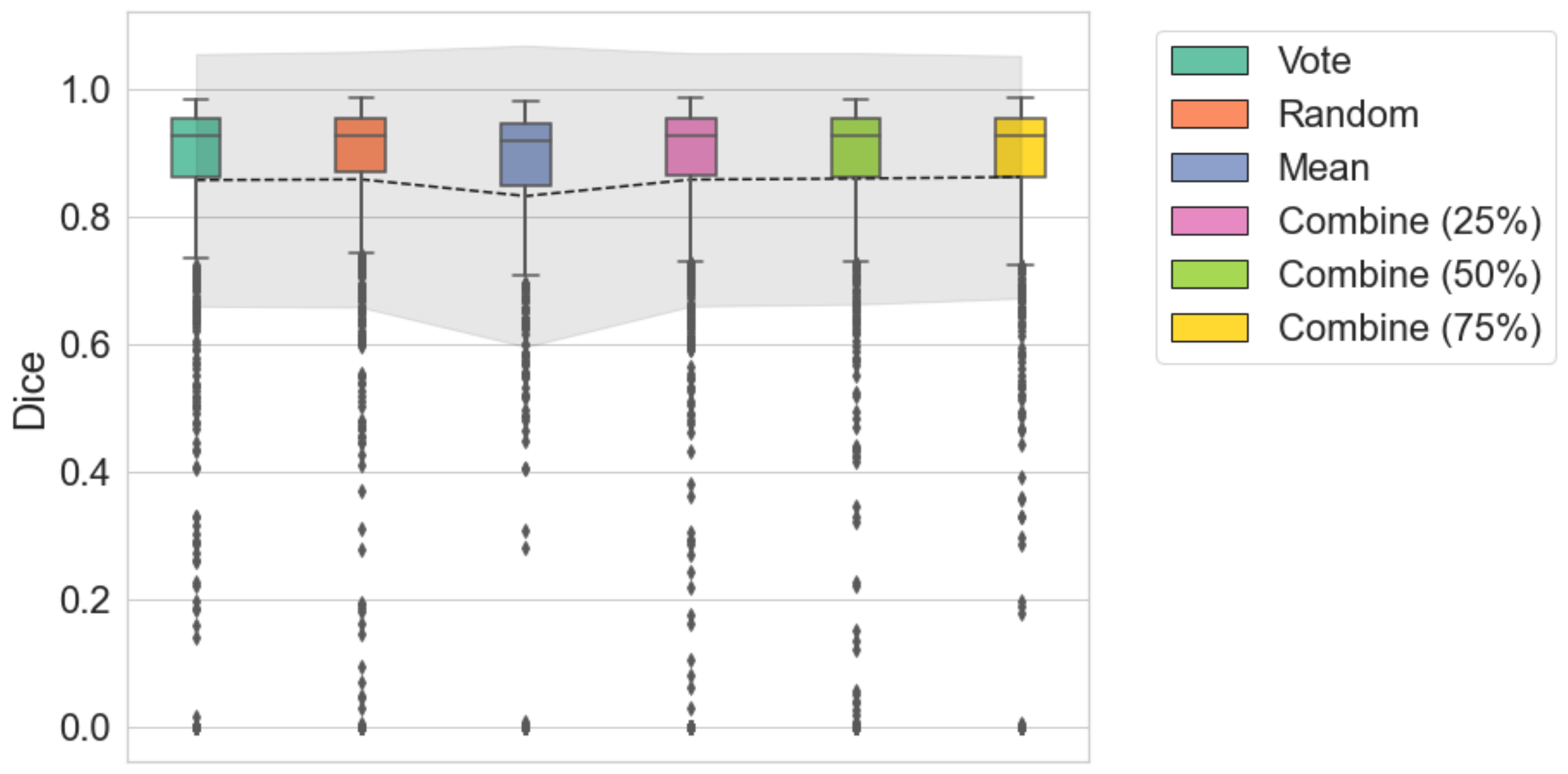}}
\hfill
\subfloat[\label{fig:mean_dice_thresh}]{\includegraphics[width=0.39\textwidth]{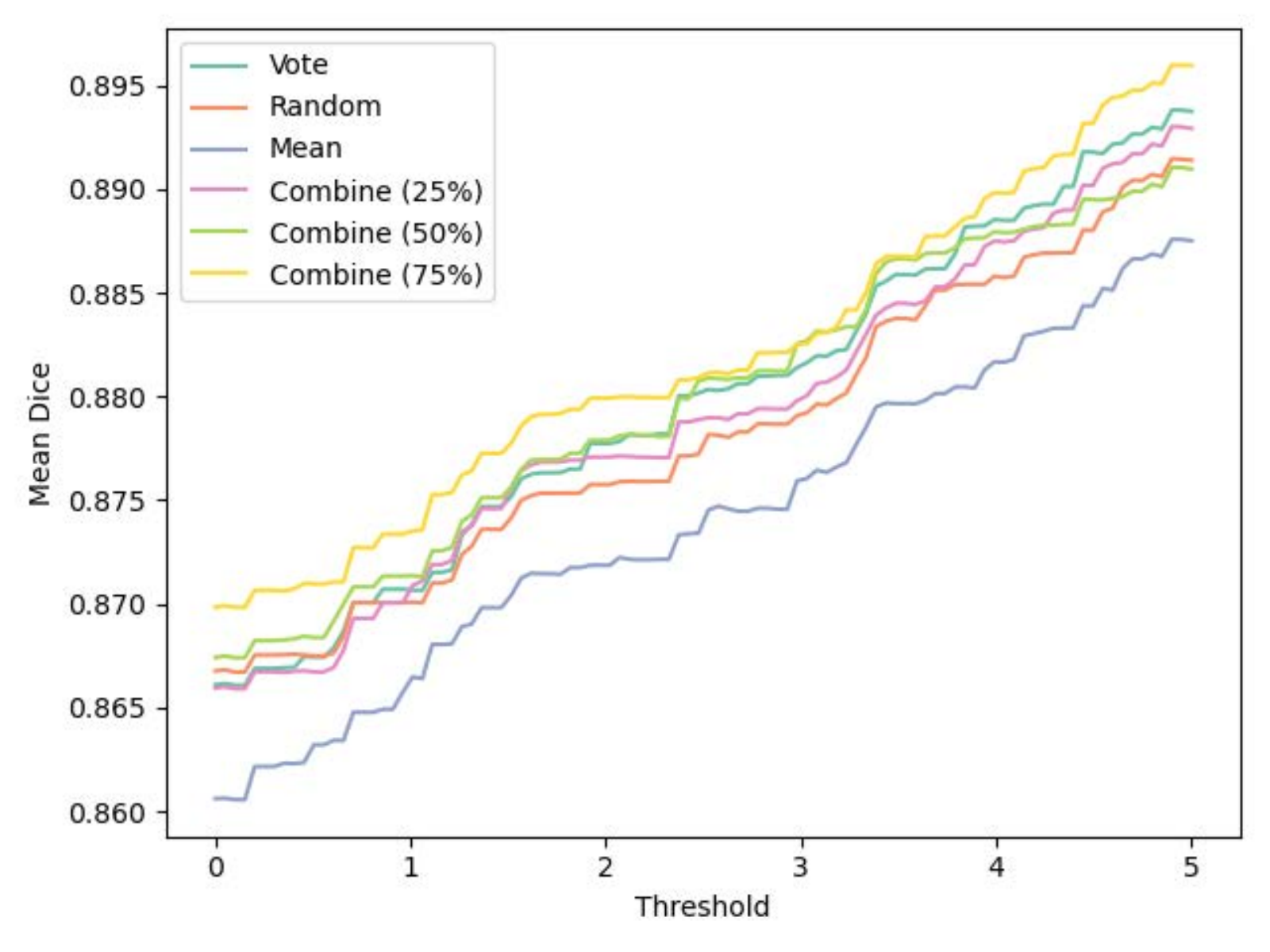}}
\caption{a) Dice coefficients for positive predictions on hold-out set of $1346$ frames. Dashed line shows mean Dice values for each strategy, with shading indicating the standard deviation from the mean. b) The mean Dice score for positive frames is reported for a range of classification thresholds for each label sampling method during segmentation. The standard deviation is omitted in the figure for readability, where for the combination strategy (25\%) we obtain a standard deviation of 0.15 at a threshold of 5.} 
\label{fig:dice_thresh_combined}
\end{figure}

The performance of the segmentation network for each sampling method is shown via box plots in Fig. \ref{fig:dice-sampling}. The mean label sampling strategy was statistically different (all $p-values$ $<$ 0.05) from all other methods. All other sampling methods obtained similar performance.

The pre-screening classifier achieved an accuracy of 97.1\% on the validation dataset during training. Table \ref{tab:results_table} summarises the Dice values with and without the pre-screening for the six label sampling methods. The classifier is shown to improve performance significantly for the mean label strategy ($p=0.001$).  

\begin{table}[h!]
\centering
\caption{The Dice coefficient values on the hold-out test data (mean ± std. dev.) with and without pre-screening. The median values are reported for inspecting skewness. Statistically significant improvement ($p<0.05$) are in bold.}\vspace{5pt}
\begin{tabular}{|c|ccc|cc|}
\hline
 Sampling & \multicolumn{3}{c|}{\bfseries Mean Dice} & \multicolumn{2}{c|}{\bfseries Median Dice} \\
\cline{2-6}
Method  & w & w/o & p-val & w & w/o \\
\hline \hline
Vote            & 0.866 ± 0.180 &  0.856 ± 0.197 & 0.220       & 0.927 & 0.926 \\
Random          & 0.867 ± 0.184 & 0.857 ± 0.200 & 0.223        & 0.926 & 0.925 \\
Mean            & \textbf{0.861 ± 0.184} & \textbf{0.831 ± 0.236}    & \textbf{0.001}    & 0.920 & 0.917 \\
Combine (25\%)  & 0.866 ± 0.182 & 0.857 ± 0.198     & 0.273    & 0.926  & 0.925 \\
Combine (50\%)  & 0.867 ± 0.180 & 0.859 ± 0.197    & 0.328    & 0.927 & 0.927 \\
Combine (75\%)  & 0.870 ± 0.174 & 0.861 ± 0.190  & 0.253      & 0.926 & 0.925 \\
\hline
\end{tabular}
\label{tab:results_table}
\end{table}

\subsubsection{Classification threshold}
The threshold used by the classifier plays a role in controlling the false positive frame rate seen by the segmentation network and can therefore be tuned as a variable at test time. We therefore tested a range of thresholds from 0 to 5 corresponding to probabilities of 0.5 to 1 and observe the effect on the mean Dice for each label sampling method. This is shown in Fig. \ref{fig:mean_dice_thresh}. From this plot the combination of consensus and random labels with a ratio of 25\% and 75\% respectively leads to the highest Dice score and this increases with threshold in general for all label sampling methods.

\subsubsection{Pre-screening classifier}
\begin{figure}[ht!]
\centering
\includegraphics[width=0.95\textwidth]{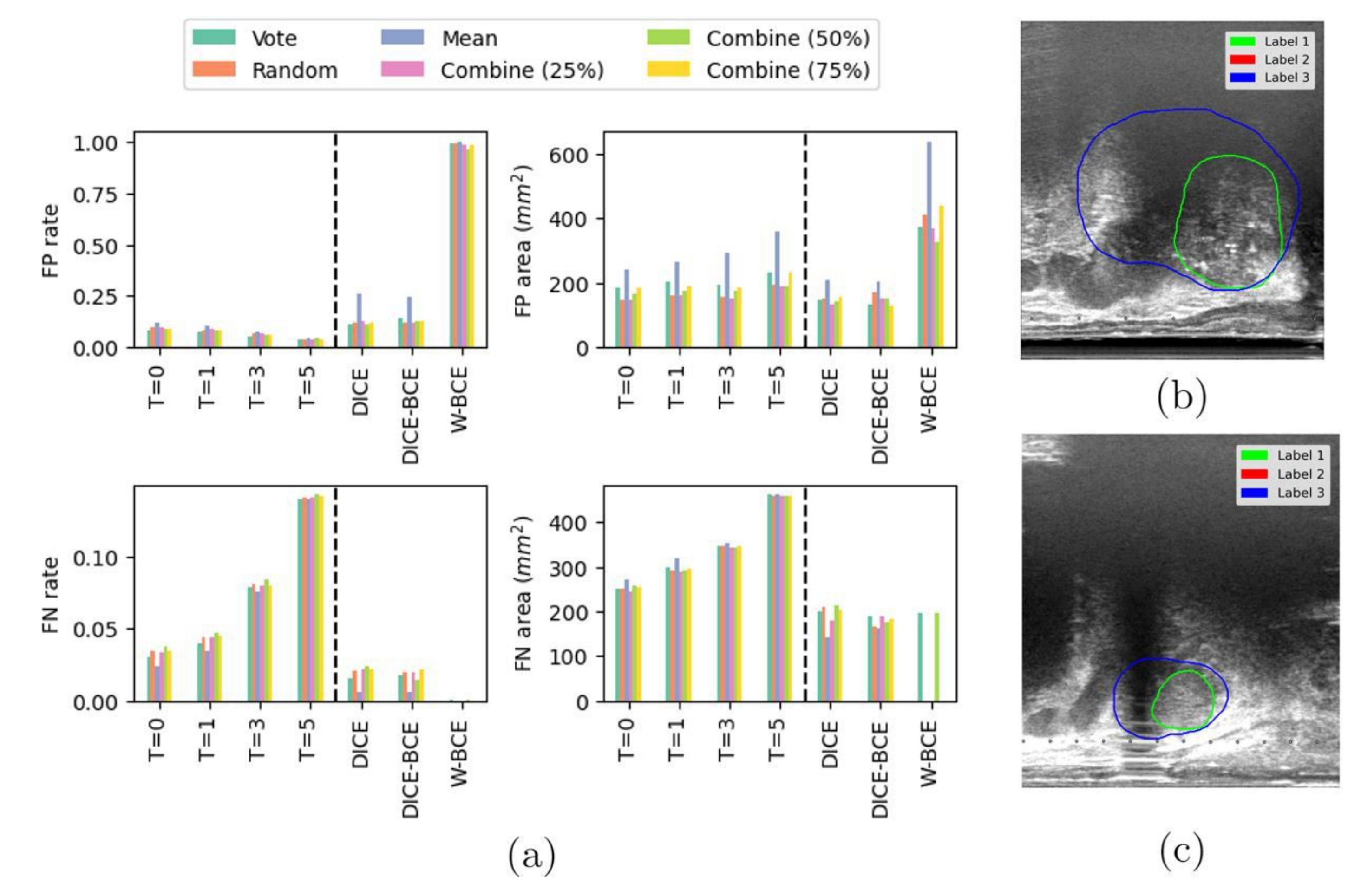}

\caption{a) False positive (FP) and false negative (FN) rates (frame-level) and areas (pixel-level) are computed for each label sampling method using different screening thresholds. We also show the rates achieved using different loss functions; Dice, Dice-BCE and W-BCE, using the segmentation-only approach (Dotted black line used to separate these cases, with classifier used in results left of line and segmentation-only results to the right)
(b-c) Example frames with manual labels, where the classifier and segmentation network disagreed. b) Classifier predicted the presence of prostate, but segmented mask is empty. c) Classifier predicted an empty frame, but the prostate was segmented. In both cases, only two labellers were in agreement, but not over the size and position of the prostate.}
\label{fig:disagree_images}
\end{figure}

The pre-screened segmentation model can be used to examine the effect on the number of false positives/negatives on both frame and pixel levels. We also use the modified loss functions to compare the performance of the segmentation network alone with a loss chosen to tackle both tasks simultaneously. The FPR and FNR is computed for the different labelling strategies in each case as shown in Fig. \ref{fig:disagree_images}a. From these results we observe a slight decrease in the number of false positive frames as the threshold increases. The most noticeable effect of threshold is on the FNR for which a larger threshold leads to a greater number of false negative frames. On the other hand, the loss function is shown to be effective to some extent at addressing the frame-level classification task. The losses seemed to lead to a lower false negatives than false positive frames, although altering the weight to control the two type of frame-level errors does not seem to be straightforward. This is consistent with what can be observed from the areas of the false segmentations.

Further inspecting the labels from three observers overlaid with those frames, on which the segmentation and classifier networks disagreed for the frame classification, as shown in the examples in Figure \ref{fig:disagree_images}b,c. Interestingly, relatively large disagreement between observers can also be found on those network-disagreed images. This may suggest a correlation between the label sampling methods and the frame classifying strategy. This is also supported by the results in Table \ref{tab:results_table}, where, for example, highest median Dice values may come from different label sampling methods, between models with and without the pre-screening strategy.

This paper reports experiment results with and without an independently-trained pre-screening classifer. Future work may investigate a classifer trained simultaneously with the segmentation network, such that the segmentation network could be optimised on representative frames that need to be segmented.

\section{Conclusion}
In this study, we investigated different strategies for handling multiple labels for intraoperative prostate gland segmentation on TRUS images. We demonstrate that disagreements between labellers affect the performance of a U-Net segmentation network due to the difficulty when defining a ground truth. Whilst there were no significant differences between the label sampling methods themselves using the Dice loss, by introducing a pre-screening strategy with a separate classifier, we show an improved segmentation accuracy by removing false positive frames. This was observed for the mean label strategy ($p=0.001<0.05$) between the mean Dice with, and without, pre-screening. Our results also agree in general with existing findings that using a combination of random and consensus labels (25\%, 75\% respectively) during training leads to better, and more stable performance with a mean Dice of $0.87 \pm 0.17$. Alternatively, the segmentation network can be trained using loss functions that aim to address the frame-level classification task in parallel with optimising the Dice score. For these models, we find a better ability to handle false negative frames than using a pre-screening classifier. However, the classifier still provides better flexibility to control the frame-level accuracy during test-time. This work illustrates the potential benefit of pre-screening prior to classification during real-time ultrasound-guided procedures where the reduction of a specific error type may be more desirable. 


\section*{Acknowledgement}
This work is supported by the EPSRC-funded UCL Centre for Doctoral Training in Intelligent, Integrated Imaging in Healthcare (i4health) (EP/S021930/1) and the Department of Health’s NIHR-funded Biomedical Research Centre at University College London Hospitals. Z.M.C. Baum is supported by the Natural Sciences and Engineering Research Council of Canada Postgraduate Scholarships-Doctoral Program, and the University College London Overseas and Graduate Research Scholarships. This work is also supported by the Wellcome/EPSRC Centre for Interventional and Surgical Sciences (203145Z/16/Z).
\bibliographystyle{splncs04}
\bibliography{bibliography}

\begin{thebibliography}{10}
\providecommand{\url}[1]{\texttt{#1}}
\providecommand{\urlprefix}{URL }
\providecommand{\doi}[1]{https://doi.org/#1}

\bibitem{Anas2018}
Anas, E.M.A., Mousavi, P., Abolmaesumi, P.: A deep learning approach for real
  time prostate segmentation in freehand ultrasound guided biopsy. Medical
  Image Analysis  \textbf{48},  107--116 (Aug 2018),
  \url{https://doi.org/10.1016/j.media.2018.05.010}

\bibitem{Ghose2012}
Ghose, S., Oliver, A., et~al.: A survey of prostate segmentation methodologies
  in ultrasound, magnetic resonance and computed tomography images. Computer
  Methods and Programs in Biomedicine  \textbf{108}(1),  262--287 (Oct 2012),
  \url{https://doi.org/10.1016/j.cmpb.2012.04.006}

\bibitem{Giordano}
Giordano, D., Kavasidis, I., et~al.: Rejecting false positives in video object
  segmentation. In: Azzopardi, G., Petkov, N. (eds.) Computer Analysis of
  Images and Patterns. pp. 100--112. Springer International Publishing, Cham
  (2015)

\bibitem{resnet}
He, K., Zhang, X., et~al.: Deep residual learning for image recognition (2015)

\bibitem{Hossain2019}
Hossain, M.S., Paplinski, A.P., Betts, J.M.: Prostate segmentation from
  ultrasound images using residual fully convolutional network (2019)

\bibitem{Kingma2017}
Kingma, D.P., Ba, J.: Adam: A method for stochastic optimization (2017)

\bibitem{Lei2019}
Lei, Y., Tian, S., et~al.: Ultrasound prostate segmentation based on
  multidirectional deeply supervised v-net. Medical Physics  \textbf{46}(7),
  3194--3206 (May 2019), \url{https://doi.org/10.1002/mp.13577}

\bibitem{Orlando2020}
Orlando, N., Gillies, D.J., et~al.: Automatic prostate segmentation using deep
  learning on clinically diverse 3d transrectal ultrasound images. Medical
  Physics  \textbf{47}(6),  2413--2426 (Apr 2020),
  \url{https://doi.org/10.1002/mp.14134}

\bibitem{unet_conv}
Ronneberger, O., Fischer, P., Brox, T.: U-net: Convolutional networks for
  biomedical image segmentation. CoRR  \textbf{abs/1505.04597} (2015),
  \url{http://arxiv.org/abs/1505.04597}

\bibitem{Rottman2019}
Rottmann, M., Maag, K., Chan, R., Hüger, F., Schlicht, P., Gottschalk, H.:
  Detection of false positive and false negative samples in semantic
  segmentation (2019)

\bibitem{Sarkar2016}
Sarkar, S., Das, S.: A review of imaging methods for prostate cancer detection.
  Biomedical Engineering and Computational Biology  \textbf{7s1},  BECB.S34255
  (Jan 2016), \url{https://doi.org/10.4137/becb.s34255}

\bibitem{Sudre2019}
Sudre, C.H., Anson, B.G., et~al.: Let's agree to disagree: learning highly
  debatable multirater labelling. CoRR  \textbf{abs/1909.01891} (2019),
  \url{http://arxiv.org/abs/1909.01891}

\bibitem{Sudre2017}
Sudre, C.H., Li, W., Vercauteren, T., Ourselin, S., Jorge~Cardoso, M.:
  Generalised dice overlap as a deep learning loss function for highly
  unbalanced segmentations. Lecture Notes in Computer Science p. 240–248
  (2017), \url{http://dx.doi.org/10.1007/978-3-319-67558-9_28}

\bibitem{Wang2018}
Wang, Y., Deng, Z., Hu, X., Zhu, L., Yang, X., Xu, X., Heng, P.A., Ni, D.: Deep
  attentional features for prostate segmentation in ultrasound. In: Medical
  Image Computing and Computer Assisted Intervention {\textendash} {MICCAI}
  2018, pp. 523--530. Springer International Publishing (2018),
  \url{https://doi.org/10.1007/978-3-030-00937-3_60}

\bibitem{Xie2017}
Xie, S., Girshick, R., Dollár, P., Tu, Z., He, K.: Aggregated residual
  transformations for deep neural networks (2017)

\end{thebibliography}

\end{document}